\documentclass[letterpaper]{article} 
\usepackage{aaai23}
\usepackage{times}  
\usepackage{helvet}  
\usepackage{courier}  
\usepackage[hyphens]{url}  
\usepackage{graphicx} 
\usepackage{natbib}  
\usepackage{caption} 
\usepackage{algorithm}
\usepackage{algorithmic}

\usepackage{textcomp}
\usepackage{amssymb}
\usepackage{multirow}
\usepackage{amsmath}
\usepackage{booktabs}
\usepackage{pifont}
\usepackage{color}
\usepackage{bm}

\usepackage{newfloat}
\usepackage{listings}
\DeclareCaptionStyle{ruled}{labelfont=normalfont,labelsep=colon,strut=off} 
\floatstyle{ruled}
\newfloat{listing}{tb}{lst}{}
\floatname{listing}{Listing}
\newcommand{{\model}}{SeqUST}
\title{Uncertainty-aware Self-training for Low-resource Neural Sequence Labeling}
\author {
    Jianing Wang \textsuperscript{\rm 1},
    Chengyu Wang \textsuperscript{\rm 2},
    Jun Huang \textsuperscript{\rm 2},
    Ming Gao \textsuperscript{\rm 1,3}\thanks{\ \ Corresponding author.},
    Aoying Zhou \textsuperscript{\rm 1}
}
\affiliations {
    \textsuperscript{\rm 1} School of Data Science and Engineering, East China Normal University, Shanghai, China \\
    \textsuperscript{\rm 2} Alibaba Group, Hangzhou, China \\
    \textsuperscript{\rm 3} KLATASDS-MOE, School of Statistics, East China Normal University, Shanghai, China \\
    lygwjn@gmail.com, \{chengyu.wcy, huangjun.hj\}@alibaba-inc.com, \{mgao, ayzhou\}@dase.ecnu.edu.cn
}

\usepackage{bibentry}

\begin{document}

\maketitle

\begin{abstract}
Neural sequence labeling (NSL) aims at assigning labels for input language tokens, which covers a broad range of applications, such as named entity recognition (NER) and slot filling, etc. 
However, the satisfying results achieved by traditional supervised-based approaches heavily depend on the large amounts of human annotation data, which may not be feasible in real-world scenarios due to data privacy and computation efficiency issues. 
This paper presents~{\model}, a novel uncertain-aware self-training framework for NSL to address the labeled data scarcity issue and to effectively utilize unlabeled data. 
Specifically, we incorporate Monte Carlo (MC) dropout in Bayesian neural network (BNN) to perform uncertainty estimation at the token level and then select reliable language tokens from unlabeled data based on the model confidence and certainty. A well-designed masked sequence labeling task with a noise-robust loss supports robust training, which aims to suppress the problem of noisy pseudo labels.
In addition, we develop a Gaussian-based consistency regularization technique to further improve the model robustness on Gaussian-distributed perturbed representations. This effectively alleviates the over-fitting dilemma originating from pseudo-labeled augmented data. 
Extensive experiments over six benchmarks demonstrate that our~{\model} framework effectively improves the performance of self-training, and consistently outperforms strong baselines by a large margin in low-resource scenarios. 
\end{abstract}

\section{Introduction}
Neural sequence labeling (NSL) is one of the fundamental tasks in natural language processing (NLP) with a broad range of applications, including named entity recognition (NER)~\cite{Li2022A} and slot filling~\cite{Liu2022Cross}, which aims at classifying language tokens into a pre-defined set of classes~\cite{Shen2021Locate, Agarwal2022Towards}. 
Previous works have achieved satisfying performance by developing well-designed deep architectures and/or fine-tuning pre-trained language models (PLMs)~\cite{Ma2016End, Li2020A, Zhou2022Distantly, Zhang2022Exploring}. Yet, these approaches heavily depend on massive labeled data, which could be even more bothersome in some real-world scenarios, such as over-fitting on limited labeled data and privacy constraints.

Recently, a branch of the semi-supervised learning (SSL) paradigm~\cite{Nitesh2005Learning, Jesper2020A, Yang2020A} arises to bypass the aforementioned issues, which aims to utilize effectively large-scale unlabeled data in addition to the few-shot labeled data. Self-training is one of the typical SSL techniques which can be viewed as teacher-student training~\cite{Grandvalet2004Semi, Amini2022Self}. Concretely, a~\emph{teacher} model is trained over the labeled data and then be used to generate pseudo labels on the unlabeled data. After that, a~\emph{student} model can be optimized by the augmented data, and be used to initialize a new~\emph{teacher}.

Benefit from self-training, previous methods make a remarkable success on a series of~\emph{instance-level classification} tasks, such as image classification~\cite{Zhou2021Instant, Wang2022Double, Liu2021Unbiased} and text classification~\cite{Meng2020Text, Mukherjee2020Uncertainty, Yu2021ATM, Hu2021Uncertainty, Tsai2022Contrast, Kim2022LST}. 
In contrast to instance-level classification, we observe that there are two challenges in applying standard self-training to NSL. On one hand, the task of NSL is based on the~\emph{token-level classification}, which requires the model to capture the inherent token-wise label dependency.
On the other hand, the teacher model inevitably generates some noisy labels that cause error accumulation~\cite{Wang2021Meta}.
Some sample selection strategies (e.g., model confidence, uncertainty estimation) and consistency regularization mitigate the effect of noisy labels and alleviate the problem of confirmation bias~\cite{Do2021Semi, Cao2021Uncertainty, Rizve2021In, Wang2021Combating, Andersen2022Efficient}. However, it is unclear how these methods can be applied to token-level classification. 

To remedy this dilemma, we develop~{\model}, a novel semi-supervised learning framework for NSL, which improves standard self-training via two decomposed processes, i.e., reliable token selection and robust learning. 
Specifically, we first pseudo annotate the unlabeled data.
Then, a Monte Carlo (MC) dropout technique, which is the approximation technique in Bayesian Neural Network (BNN)~\cite{Gal2016Dropout, Wang2016Towards}, is used to estimate the uncertainty of each language token derived from the teacher model.
We judiciously select the reliable tokens from each unlabeled sentence based on the model confidence and certainty. 
Finally, we introduce two training objectives to improve the model robustness, i.e., Masked Sequence Labeling (MSL) and Gaussian-based Consistency Regularization (GCR).
In MSL, we generate a masked matrix to make the model only focus on the selected reliable tokens in each sentence. We also utilize partially huberised cross-entropy (PHCE) loss to explicitly mitigate the effect of label noises.
In GCR, we assume each selected token embedding follows the Gaussian distribution and perturbs token embeddings to encourage consistency between perturbed embeddings and original representations. This technique effectively reduces the risk of over-fitting.

We perform extensive experiments over multiple benchmarks with very few labeled data (10/20/50/100 examples per class). We adopt BERT-base~\cite{Devlin2019BERT} as our backbone. Results show that our~{\model} outperforms strong baselines by a large margin in low-resource settings.

In a nutshell, we make the following main contributions:
\begin{itemize}
    \item We develop a novel semi-supervised neural sequence labeling framework~{\model} to alleviate the problem of data scarcity in low-resource scenarios.
    \item We propose the token-level uncertainty estimation to empower self-training. Furthermore, two training objectives are introduced to improve the model robustness, including Masked Sequence Labeling and Gaussian-based Consistency Regularization.
    \item Extensive experiments over multiple benchmark datasets demonstrate that our proposed framework achieves substantial performance improvement.
\end{itemize}

\section{Related Work}

\subsection{Semi-supervised Learning and Self-training}
SSL aims to utilize effectively unlabeled data in addition to labeled data, which has been widely used in the NLP community~\cite{Yang2017Improved, Gururangan2019Variational, Xie2020Unsupervised, Chen2020MixText}. 
For instance, \citet{Yang2017Improved, Gururangan2019Variational} utilize variational autoencoders (VAEs) for sequence classification and labeling.
Unsupervised data augmentation (UDA)~\cite{Xie2020Unsupervised} generates augmented data by back translation, and leverages consistency regularization for unlabeled data.
\citet{Chen2020MixText} proposes {MixText} to mix labeled, unlabeled and augmented data, and performs similar consistency training as UDA.

Self-training is one of the earliest methods in SSL that has recently shown state-of-the-art performances~\cite{Chen2020MixText, Meng2020Text, Li2021Semi}. Recent works improve standard self-training by considering sample selection strategies, including model confidence~\cite{Bengio2009Curriculum, Kumar2010Self} and uncertainty estimation~\cite{Cao2021Uncertainty, Tsai2022Contrast}. 
For example, \citet{Cao2021Uncertainty} presents uncertainty-aware self-training (UST) to sample pseudo-labeled data by BNN. ~\cite{Tsai2022Contrast} propose CEST which aims to leverage graph-based contrast induction to solve the problem of confirmation bias and data smoothness on the selected data. 
However, these approaches mainly focus on instance-level classification. It would be highly desirable if they can be applied to token-level classification.

\subsection{Low-resource Neural Sequence Labeling}
Low-resource NSL aims at classifying the input language tokens with very few labeled data. Prior researches address this problem via meta-learning~\cite{Ziyadi2020Example, Ma2022Decomposed}
and augmentation learning~\cite{Ding2020DAGA, Zhou2022MELM, Wang2022PromDA}. For instance, \cite{Ziyadi2020Example} leverages prototypical network~\cite{Snell2017Prototypical} to learn adaptable knowledge from few-shot episode data. 
\citet{Wang2022PromDA} and \citet{Zhou2022MELM} aim to generate the in-domain augmented data through masked language modeling (MLM)~\cite{Devlin2019BERT}.

Yet, these methods ignore the informative semantics of unlabeled data. To reach this goal, a series of previous works~\cite{Miller2004Name, Peters2017Semi} focus on token representations enhancement by pre-training word embeddings on unlabeled data.
Another line of research focuses on latent variable modeling~\cite{Chen2018Variational}, adversarial training method SeqVAT~\cite{Chen2020SeqVAT} and cross-view training method CVT~\cite{Clark2018Semi}.
Recently, MetaST~\cite{Wang2021Meta} attempts to apply self-training to NSL by two techniques, i.e., adaptive validation set construction by uncertainty estimation, and noisy data re-weighting via student loss.
In contrast, we perform reliable token selection by the joint estimations of the teacher confidence and uncertainty, and two well-designed training objectives support robust training. 
Compared with previous methods, we achieve the best overall performance with very few labeled data.

\begin{figure*}[t]
\centering
\includegraphics[width=\linewidth]{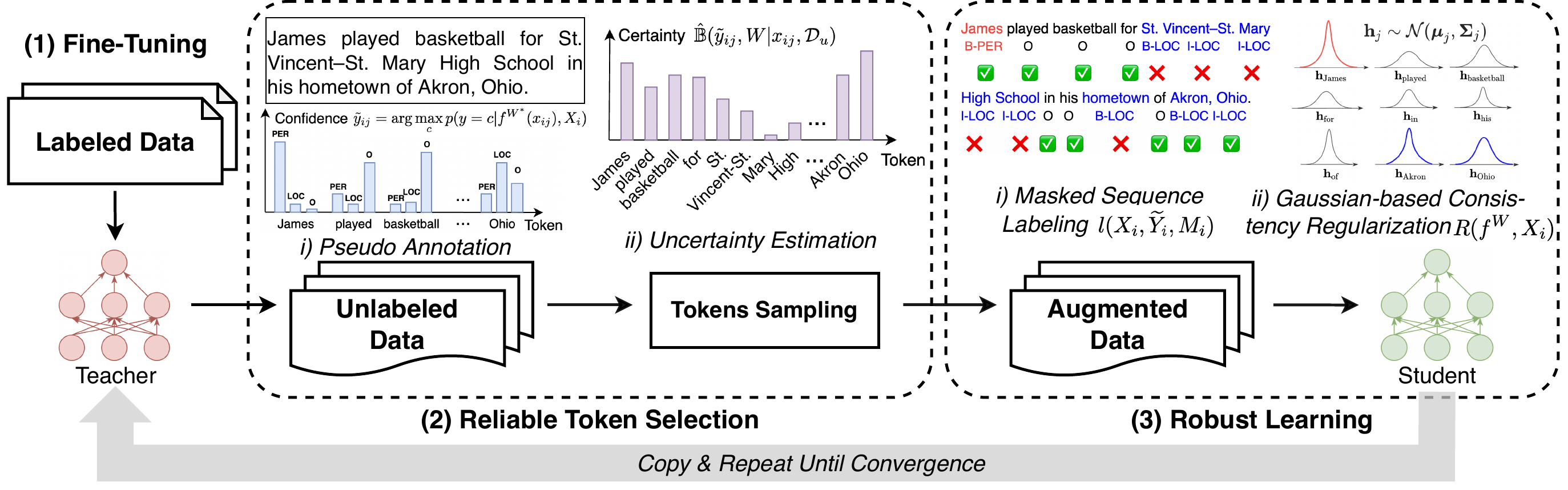} 
\caption{The framework overview. 1) We first fine-tune a teacher model over the labeled data. Then, 2) we utilize MC dropout to perform uncertainty estimation and select reliable tokens for each unlabeled sentence. 3) Two novel training objectives are proposed to improve the model robustness and alleviate the over-fitting issue. (Best viewed in color.)}.
\label{fig:model}
\end{figure*}

\section{Preliminaries}
We first present the notations and then introduce the background knowledge of the Bayesian neural network.

\subsection{Notations}
We represent a labeled dataset and an unlabeled dataset as $\mathcal{D}_{l}=\{X_i, Y_i\}_{i=1}^{N_l}$ and $\mathcal{D}_{u}=\{X_i\}_{i=1}^{N_u}$, respectively. $X_i=\{x_{ij}\}_{j=1}^{L}$ denotes the input sentence with $L$ language tokens. $Y_i=\{y_{ij}\}_{j=1}^{L}$ is the label sequence and $y_{ij}\in\mathcal{Y}$ is the tag of token $x_{ij}\in\mathcal{X}$, $\mathcal{X}$ and $\mathcal{Y}$ denote the token vocabulary of the PLM and the label space, respectively. $N_l$ and $N_u$ represent the numbers of labeled and unlabeled data, respectively ($N_l\ll N_u$).
The goal of semi-supervised NSL is to learn a neural mapping function $f^{W}:\mathcal{X}\rightarrow\mathcal{Y}$ over labeled data $\mathcal{D}_{l}$ and unlabeled data $\mathcal{D}_{u}$, where $W$ is the collection of model parameters.

\subsection{Bayesian Neural Network (BNN)}
Similar to CEST~\cite{Tsai2022Contrast}, in this part we briefly describe BNN.
Given a neural model $f^{W}$, the vanilla BNN assumes a prior distribution over its model parameters $W$. In other words, BNN averages over all the possible weights instead of directly optimizing for the weights~\cite{Mukherjee2020Uncertainty}. 
Given a labeled dataset $\mathcal{D}_{l}$, the parameter can be optimized by the posterior distribution $p(W|\mathcal{D}_{l})$. 
During model inference, given one unlabeled example $X_i\in\mathcal{D}_{u}$, the probability distribution of each token $x_{ij}\in X_i$ for class $c\in\mathcal{Y}$ is $p(y=c|x_{ij}, X_i)=\int_{W}p(y=c|f^{W}(x_{ij}, X_i)p(W|D_{u})dW$. 
Yet, it is intractable in practice for calculation. 
To make it tractable, we can find a surrogate tractable distribution $q_{\theta}(W)$ that makes the model posterior easy to compute. 
Thus, we consider $q_{\theta}(W)$ to be the dropout distribution~\cite{Srivastava2014Dropout} which aims to sample $T$ masked model weights $\{\widetilde{W}_t\}_{t=1}^{T}\sim q_{\theta}(W)$ from the current model. The approximate posterior for each token is:
\begin{equation}
\begin{aligned}
p(y=c|x_{ij}, X_i)\approx\frac{1}{T}\sum_{t=1}^{T}p(y=c|f^{\widetilde{W}_t}(x_{ij}), X_i).
\label{eql:bnn-posterior}
\end{aligned}
\end{equation}

\section{Methodology}

In this section, we propose the~{\model} framework to improve the self-training paradigm for low-resource NSL.
The framework overview is illustrated in Fig~\ref{fig:model}.




\subsection{Pseudo Annotation}
In the initial stage, a vanilla PLM (e.g., BERT)
can be fine-tuned over the labeled dataset $\mathcal{D}_{l}$ to form a \emph{teacher} model $f^{W^*}$, where $W^*$ is the collection of parameters. 
The hard label $\tilde{y}_{ij}$ of each token $x_{ij}$ in the given unlabeled sentence $X_i\in\mathcal{D}_{u}$ can be pseudo annotated by the teacher model $f^{W^*}$. Formally, we have:
\begin{equation}
\begin{aligned}
\tilde{y}_{ij}=\arg\max_{c}p(y=c|f^{W^{*}}(x_{ij}), X_i),
\label{eql:pseudo-label}
\end{aligned}
\end{equation}
where $p(\cdot)$ is the probability distribution, which can be modeled as a softmax classifier or a conditional random field (CRF)~\cite{Lafferty2001Conditional} layer. 


\subsection{Reliable Token Selection}


Prior works~\cite{Mukherjee2020Uncertainty, Tsai2022Contrast, Rizve2021In} improve self-training by the instance-level selection strategy, which aims to utilize BNN to perform uncertainty estimation for each example, and then select the reliable examples from the whole unlabeled dataset that the model is most certain about.
Different from them, we focus on the token-level uncertainty estimation, and aim to select reliable tokens from each sentence. 

\noindent\textbf{Token-level Uncertainty Estimation}.
We assume that each sentence is independent of another and can be measured individually.
Specifically, we follow~\cite{Houlsby2011Bayesian, Gal2017Deep} to leverage information gain of the model parameters to estimate how certain the model is to the pseudo-labeled tokens with respect to the true labels.
Formally, given one input sentence $X_i\in\mathcal{D}_{u}$, we have:
\begin{equation}
\begin{aligned}
\mathbb{B}(\tilde{y}_{ij}, W|x_{ij}, \mathcal{D}_{u}) = 
& \mathbb{H}(\tilde{y}_{ij}|x_{ij}, \mathcal{D}_{u}) - \\
& \mathbb{E}_{p(W|\mathcal{D}_{u})}[\mathbb{H}(\tilde{y}_{ij}|x_{ij}, W)],
\label{eql:information-gain}
\end{aligned}
\end{equation}
where $\mathbb{H}(\cdot)$ is the entropy function, $x_{ij}\in X_i$ and $\tilde{y}_{ij}\in\widetilde{Y}_i$ denote the token and tag, respectively. 
$\mathbb{B}(\tilde{y}_{ij}, W|x_{ij}, \mathcal{D}_{u})$ denotes the information gain which is the difference between $\mathbb{H}(\tilde{y}_{ij}|x_{ij}, \mathcal{D}_{u})$ (the final entropy after seeing all tokens from unlabeled sentences) and $\mathbb{H}(\tilde{y}_{ij}|x_{ij}, W)$ (the current entropy for the token $x_{ij}$).
$p(W|\mathcal{D}_{u})$ is the posterior distribution.
As the calculation of Eq.~\ref{eql:information-gain} is intractable, we utilize MC dropout in BNN to perform approximation. Specifically, we assume that the posterior distribution $p(W|\mathcal{D}_{u})$ can be replaced with dropout distribution $q_{\theta}(W)$. Thus, we can sample $T$ masked model weight $\{\widetilde{W}_t\}_{t=1}^{T}\sim q_{\theta}(W)$, and calculate the approximation value of $\mathbb{B}$ as:
\begin{equation}
\begin{aligned}
\hat{\mathbb{B}}(\tilde{y}_{ij}, W|x_{ij}, \mathcal{D}_{u}) = 
&- \sum_{c\in\mathcal{Y}}(\frac{1}{T}\sum_{t=1}^{T}\hat{p}_c^t)\log(\frac{1}{T}\sum_{t=1}^{T}\hat{p}_c^t) \\
&+ \frac{1}{T}\sum_{t=1}^{T}\sum_{c\in\mathcal{Y}}\hat{p}_c^t\log(\hat{p}_c^t),
\label{eql:information-gain-mc}
\end{aligned}
\end{equation}
where $\hat{p}_c^t=p(\tilde{y}_{ij}=c|f^{\widetilde{W}_t}(x_{ij}), X_i)$ is the predict probability for the token $x_{ij}$ derived from the $t$-th masked model.
Thus, a lower $\hat{\mathbb{B}}(\tilde{y}_{ij}, W|x_{ij}, \mathcal{D}_{u})$ value means that the model is more certain about the prediction, as higher certainty corresponds to lower information gain.

\noindent\textbf{Tokens Sampling}.
For reliable token selection, we jointly consider the confidence and certainty value.
For the confidence of the prediction $\tilde{y}_{ij}$, we have:
\begin{equation}
\begin{aligned}
s_{ij}^{cf}=\frac{1}{T}\sum_{t=1}^{T}p(y=\tilde{y}_{ij}|f^{\widetilde{W}_t}(x_{ij}), X_i).
\label{eql:confidence-score}
\end{aligned}
\end{equation}
A higher confidence value $s_{ij}^{cf}$ means the model is more confident for the pseudo label $\tilde{y}_{ij}$. 
Theoretically, selecting language tokens with high confidence predictions moves decision boundaries to low-density regions, which is satisfied with the low-density assumption~\cite{Rizve2021In}.
However, many of these selected tokens with higher confidence are incorrect due to the poor calibration of neural networks~\cite{Guo2017On}, which brings the conformation bias problem.
To reduce the wrong labels, we additionally design a certainty score $s_{ij}^{ct}$ based on the uncertainty estimation as:
\begin{equation}
\begin{aligned}
s_{ij}^{ct}=1 - \hat{\mathbb{B}}(\tilde{y}_{ij}, W|x_{ij}. \mathcal{D}_{u}),
\label{eql:certainty-score}
\end{aligned}
\end{equation}
Intuitively, if the model is always certain about some tokens, these tokens might be too easy to contribute any additional information.
To this end, we can obtain the final sampling weight for each token as:
\begin{equation}
\begin{aligned}
s_{ij}=\frac{s_{ij}^{cf}\times s_{ij}^{ct}}{\sum_{x_{ij}\in X_i}s_{ij}^{cf}\times s_{ij}^{ct}}.
\label{eql:sampling-weight}
\end{aligned}
\end{equation}



\subsection{Robust Learning for NSL}
\noindent\textbf{Masked Sequence Labeling (MSL).}
With the measure of model confidence and certainty, we can use them to sample reliable tokens for each sentence. 
However, each token has the location and label dependency constraint, we can not directly remove the tokens that are not sampled. 
Thus, we define a masked matrix to record the sampling results, i.e., 
\begin{equation}
\begin{aligned}
M_{i, j} = \left\{
        \begin{array}{rcl}
        1     & x_{ij}~\text{has been sampled}; \\
        0     & x_{ij}~\text{has not been sampled}; \\
        \end{array} 
    \right.
\label{eql:visible-matrix}
\end{aligned}
\end{equation}
when $M_{ij}=0$, it means the corresponding label may be a noise and should be masked during self-training.

Generally, we can define the following cross-entropy function as the training loss:
\begin{equation}
\begin{aligned}
l(X_i, \widetilde{Y}_{i}, M_i)=\frac{1}{L_i'}\sum_{j=1}^{L_i'}\mathbb{I}(M_{ij}=1)\log p_{W}(x_{ij}, \tilde{y}_{ij}),
\label{eql:loss-cross-entropy}
\end{aligned}
\end{equation}
where $p_{W}(x_{ij}, \tilde{y}_{ij})=p(y=\tilde{y}_{ij}|f^{W}(x_{ij}), X_i)$ is the prediction probability derived from the student model, $L_i'=\sum_{j=1}^{L}\mathbb{I}(M_{ij}=1)$ is the number of the selected tokens in $X_i$. $\mathbb{I}(\cdot)$ denotes the indicator function.
However, it is still possible that the selected tokens could be wrong pseudo annotated although the token with higher confidence and certainty.
To explicitly mitigate the effect of label noises, we follow~\cite{Tsai2022Contrast} to utilize partially huberised cross-entropy loss (PHCE loss) as our noise-robust loss, which is based on a simple variant of gradient clipping for the classification loss, e.g. cross-entropy. 
Hence, the loss function in Eq.~\ref{eql:loss-cross-entropy} can be modified as:
\begin{equation}
\begin{aligned}
l(X_i, \widetilde{Y}_{i}, M_i)=\frac{1}{L_i'}\sum_{j=1}^{L_i'}\mathbb{I}(M_{ij}=1)\phi(x_{ij}, \tilde{y}_{ij}),
\label{eql:loss-1}
\end{aligned}
\end{equation}
where
\begin{equation}
\begin{aligned}
\phi(x, y) = \left\{
        \begin{array}{rcl}
        -\tau p_{W}(x, y)  
        + \log\tau + 1  & p_{W}(x, y)\leq 1/\tau; \\
        -\log p_{W}(x, y)     & p_{W}(x, y)>1/\tau; \\
        \end{array} 
    \right.
\label{eql:loss-phce}
\end{aligned}
\end{equation}
is the PHCE loss function, $\tau>1$ is the hyper-parameter. Thus, the model learned by Eq.~\ref{eql:loss-phce} can be more robust to the noisy labeled tokens than the common cross-entropy.

\begin{algorithm}[t]
\caption{Self-training Procedure of {\model}}
\label{alg:train}
\begin{small}
\begin{algorithmic}[1]
\REQUIRE Neural model $f^{W_{0}}$, labeled data $\mathcal{D}_{l}$, unlabeled data $\mathcal{D}_{u}$.
\STATE Initialize a teacher model $f^{W^{*}}=f^{W_0}$;
\WHILE{not converged}
\STATE Fine-tune the teacher model $f^{W^{*}}$ over the labeled data $\mathcal{D}_{l}$;
\STATE Pseudo annotate each unlabeled sentence $X_i\in\mathcal{D}_{u}$ by Eq.~\ref{eql:pseudo-label} to obtain the hard labels $\widetilde{Y}_i$;
\STATE Initialize a student model $f^{W}=f^{W_{0}}$;
\FOR{$X_i\subseteq\mathcal{D}_{u}$}
\STATE Obtain the confidence score $s_{ij}^{cf}$ for $x_{ij}\in X_i$ by Eq.~\ref{eql:confidence-score};
\STATE Obtain the certainty score $s_{ij}^{ct}$ for $x_{ij}\in X_i$ by Eq.~\ref{eql:certainty-score};
\STATE Sample reliable tokens by the sampling weight in Eq.~\ref{eql:sampling-weight}, and generate a masked matrix $M_{i}$;
\STATE Calculate the training loss $l(X_{i}, \widetilde{Y}_{i}, M_{i})$ in Eq.~\ref{eql:loss-1};
\STATE Calculate the regularization loss $R(f^{W}, X_i)$ in Eq.~\ref{eql:loss-regularization};
\STATE Update the model $f^{W}$ by reduce $\mathcal{L}(W)$ in Eq.~\ref{eql:loss};
\ENDFOR
\STATE Update the teacher model $f^{W^{*}}=f^{W}$;
\ENDWHILE
\RETURN The teacher model $f^{W^{*}}$.
\end{algorithmic}
\end{small}
\end{algorithm}

\noindent\textbf{Gaussian-based Consistency Regularization (GCR).}
During iterative self-training, it is possible that the model biases the sampling process toward picking \emph{easier} tokens, which have higher confidence and certainty. This inevitably leads to the student model over-fitting on these frequently selected samples.
Previous methods~\cite{Chen2020MixText, Xie2020Unsupervised} solve this problem by utilizing back translation~\cite{Sennrich2016Sennrich} to generate instance-level augmented data, which aims to translate each sentence into a different language and then translate it back. However, it can not be directly applied to token-level task~\cite{Lee2021Learning}. 
To reach this goal, we elaborately design Gaussian-based consistency regularization (GCR) to perturb token embeddings in the hidden space and make the model prediction consistent.

Given one pseudo-labeled sentence $X_i=\{x_{ij}\}_{j=1}^{L}$, we can obtain the corresponding contextual representations $\mathbf{H}_i=\{\mathbf{h}_{ij}\}_{j=1}^{L}$ derived from the final hidden layer output of the student model (i.e., BERT). $\mathbf{h}_{ij}\in\mathbb{R}^{h}$ denotes the representations of the token $x_{ij}$, where $h$ is the hidden size.
We assume that token embeddings follow Gaussian distributions~\cite{Das2022CONTaiNER, Lee2021Learning}, i.e., $\mathbf{h}_{ij}\sim\mathcal{N}(\bm{\mu}_{ij}, \bm{\Sigma}_{ij})$. Specifically, we use two projection network $\mathcal{F}_{\mu}(\cdot)$ and $\mathcal{F}_{\Sigma}(\cdot)$ with ReLU activation to produce Gaussian distribution $\bm{\mu}_{ij}$ and $\bm{\Sigma}_{ij}$ for each token $x_{ij}$. Formally, we have:
\begin{equation}
\begin{aligned}
\bm{\mu}_{ij} = \mathcal{F}_{\mu}(\mathbf{h}_{ij}),
\bm{\Sigma}_{ij} = \mathcal{F}_{\Sigma}(\mathbf{h}_{ij}),
\label{eql:gaussian-parameters}
\end{aligned}
\end{equation}
where $\bm{\mu}_{ij}\in\mathbb{R}^{h}$, $\bm{\Sigma}_{ij}\in\mathbb{R}^{h}$ represent mean and diagonal covariance of the Gaussian embedding, respectively. 
We leverage the reparameterization trick~\cite{Kingma2014Auto} to perturb the input $x_{ij}$ with the sampled noise without altering its semantics. Specifically, we generate $K$ perturbed representations $\{\hat{\mathbf{h}}_{ij}^{(k)}\}_{k=1}^{K}$, where
\begin{equation}
\begin{aligned}
\hat{\mathbf{h}}_{ij}^{(k)} = \mathbf{h}_{ij}\odot(\bm{\mu}_{ij} + \bm{\Sigma}_{ij}\odot\epsilon^{(k)}).
\label{eql:gaussian-perturbation}
\end{aligned}
\end{equation}
Here, $\epsilon^{(k)}\sim\mathcal{N}(\mathbf{0}, \mathbf{I}_{h})$, and $\mathbf{0}\in\mathbb{R}^{h}$, $\mathbf{I}_{h}\in\mathbb{R}^{h}$ are the vector with zeros and identity matrix, respectively. $\odot$ denotes the element-wise multiplication. 
Afterward, the KL divergence objective can be used to control the probability distribution consistency between the original semantic representation and the perturbations for each token of $x_{ij}$:
\begin{equation}
\begin{aligned}
R(f^{W}, X_{i}) = \frac{1}{L\cdot K}\sum_{j=1}^{L}\sum_{k=1}^{K}D_{\text{KL}}\big(p_{W}(y|\mathbf{h}_{ij})|| p_{W}(y|\hat{\mathbf{h}}_{ij}^{(k)})\big)
\label{eql:loss-regularization}
\end{aligned}
\end{equation}
where $D_{\text{KL}}(\cdot||\cdot)$ denotes the KL divergence function, and $p_{W}(y|\cdot)$ is the probability distribution derived from the student model.
Finally, we update the student model by modifying the following objective:
\begin{equation}
\begin{aligned}
\mathcal{L}(W) = \sum_{X_i\in\mathcal{D}_u}\big(l(X_i, \widetilde{Y}_i, M_i) + \lambda R(f^{W}, X_i)\big),
\label{eql:loss}
\end{aligned}
\end{equation}
where $\lambda$ is the pre-defined hyper-parameter balancing the regularization term.

\subsection{Self-training Procedure}
The training algorithm of our~{\model} framework is shown in Algorithm~\ref{alg:train}. Specifically, at each iteration stage, we fine-tune a teacher model $f^{W^{*}}$ on the few-shot labeled data $\mathcal{D}_{l}$, and pseudo annotate the unlabeled data $\mathcal{D}_{u}$ (Algorithm~\ref{alg:train}, Line 3-4). Then, we select reliable tokens in each sentence by the model confidence and certainty (Algorithm~\ref{alg:train}, Line 7-9), and use them to update the parameters of the student model $f^{W}$ by modifying the final loss $\mathcal{L}(W)$ (Algorithm~\ref{alg:train}, Line 5, 10-12).
Finally, the trained student model $f^{W}$ is used to initialize a new teacher $f^{W^{*}}$, and repeat the above steps till convergence.

\section{Experiments}
\subsection{Datasets and Implementation Details}
We choose six widely used benchmarks to evaluate our~{\model} framework, including SNIPS~\footnote{\url{https://github.com/sonos/nlu-benchmark/tree/master/2017-06-custom-intent-engines}.}~\cite{Coucke2018Snips} and Multiwoz~\footnote{\url{https://github.com/budzianowski/multiwoz}.}~\cite{Budzianowski2018MultiWOZ} for slot filing, MIT Movie~\footnote{\url{https://groups.csail.mit.edu/sls/downloads/movie/}.}~\cite{Liu2013Query}, MIT Restaurant~\footnote{\url{https://groups.csail.mit.edu/sls/downloads/restaurant/}.}~\cite{Liu2013Asgard}, CoNLL-03~\cite{Sang2003Introduction} and OntoNotes~\cite{weischedel2013ontonotes} for NER. The statistics of each dataset are shown in Table~\ref{tab:datasets}.
For each dataset, we use a greedy-based sampling strategy to randomly select 10-shot labeled data per class for the few-shot labeled training set and validation set, while the remaining data are viewed as unlabeled data.

During self-training, the teacher and student model share the same model architecture. In default, we choose BERT-base-uncased~\cite{Devlin2019BERT} from HuggingFace\footnote{\url{https://huggingface.co/transformers}.} 
with a softmax layer as the base encoder. We use grid search to search the hyper-parameters.
We select five different random seeds for the dataset split and training settings among $\{12, 21, 42, 87, 100\}$.
We report the averaged F1 scores with the standard deviation on the whole testing set.
We implement our framework in Pytorch 1.8 and use NVIDIA V100 GPUs for experiments.

\subsection{Baselines}

We adopt several state-of-the-art semi-supervised methods as our strong baselines~\footnote{We do not consider $N$-way $K$-shot few-shot NSL baselines because they have different learning settings with ours.}. 
\textbf{Standard self-training (SST)}~\cite{Huang2021Few} is the simple method that generates hard pseudo labels and uses them to train a student model.
\textbf{VAT}~\cite{Miyato2019Virtual} and \textbf{SeqVAT}~\cite{Chen2020SeqVAT} utilize adversarial training method with consistency learning to improve the robustness.
\textbf{CVT}~\cite{Clark2018Semi} is based on cross-view training for semi-supervised sequence labeling.
\textbf{MetaST}~\cite{Wang2021Meta} aims to select reliable validation data and trains the student model by re-weighting strategy.
We also choose the standard \textbf{Fine-tuning} as the supervised learning-based baselines.
We reproduce their results with the same settings.

\begin{table}
\centering
\resizebox{\linewidth}{!}{
\begin{small}
\begin{tabular}{l | ccccc}
\toprule
\bf Datasets &\bf Domain &\bf Type &\bf \#Class &\bf  \#Train & \bf \#Test \\
\midrule
SNIPS & Dialogue & Slot Filling & 54  & 13.6k  & 0.7k \\
Multiwoz & Dialogue & Slot Filling & 15  & 20.3k  & 2.8k \\
MIT Movie & Review & NER & 13 & 7.8k & 2.0k \\
MIT Restaurant & Review & NER & 9 & 7.7k & 1.5k \\
CoNLL-03 & News & NER & 5 & 14.0k & 3.5k \\
OntoNotes & General & NER & 19 & 60.0k & 8.3k \\
\bottomrule
\end{tabular}
\end{small}
}
\caption{The statistics of each dataset.}
\label{tab:datasets}
\end{table}

\begin{table*}[t]
\centering
\resizebox{\linewidth}{!}{
\begin{tabular}{lcccccccc}
\toprule
\bf Baselines & \bf  SNIPS & \bf  Multiwoz & \bf  MIT Movie &  \bf  MIT Restaurant & \bf  CoNLL-03 & \bf  OntoNotes & \bf  Avg. & $\Delta$ \\
\midrule
\multicolumn{9}{l}{\textit{\textbf{Full Data}}}\\
Fine-tuning & 97.00 & 88.00 & 87.30 & 79.00 & 91.90 & 89.20 & 88.73 & - \\
\midrule
\multicolumn{9}{l}{\textit{\textbf{Few Labeled Data (10-shot)}}}\\
Fine-tuning & 79.10\textit{\textpm0.38} & 71.40\small{\textpm0.25} & 69.50\small{\textpm0.40} & 53.75\small{\textpm0.19} & 71.82\small{\textpm0.28} & 73.15\small{\textpm0.25} & 69.79 & +0.00 \\
\midrule
\multicolumn{9}{l}{\textit{\textbf{Few Labeled Data (10-shot) + Unlabeled Data}}}\\
SST & 81.07\small{\textpm0.40} & 72.25\small{\textpm0.11} & 71.14\small{\textpm0.22} & 55.13\small{\textpm0.29} & 72.86\small{\textpm0.34} & 75.07\small{\textpm0.18} & 71.25 & +1.46\\
VAT  & 79.70\small{\textpm0.22} & 72.08\small{\textpm0.30} & 68.80\small{\textpm0.39} & 54.26\small{\textpm0.33} & 72.60\small{\textpm0.33} & 73.38\small{\textpm0.27} & 70.14 & +0.35 \\
SeqVAT  & 79.62\small{\textpm0.41} & 72.17\small{\textpm0.34} & 68.83\small{\textpm0.35} & 54.02\small{\textpm0.40} & 73.18\small{\textpm0.28} & 73.71\small{\textpm0.30} & 70.23 & +0.44\\
CVT  & 79.22\small{\textpm0.44} & 71.42\small{\textpm0.38} & 69.03\small{\textpm0.51} & 54.17\small{\textpm0.42} & 71.70\small{\textpm0.72} & 73.88\small{\textpm0.40} & 69.90 & +0.11\\
MetaST  & \underline{86.74\small{\textpm0.33}} & \underline{77.34\small{\textpm0.51}} & \underline{77.52\small{\textpm0.39}} & \underline{63.02\small{\textpm0.29}} & \underline{76.88\small{\textpm0.41}} & \underline{77.69\small{\textpm0.24}} & \underline{76.53} & \underline{+6.74}\\
\midrule
\textbf{SeqUST}  & \bf 87.33\small{\textpm0.30} & \bf 77.98\small{\textpm0.26} & \bf 77.73\small{\textpm0.22} & \bf 64.19\small{\textpm0.36} & \bf 79.10\small{\textpm0.27} & \bf 80.33\small{\textpm0.44} & \bf 77.78 & \bf +7.99\\
\bottomrule
\end{tabular}
}
\caption{The performance comparison of F1 scores (\%) with standard deviations on six benchmarks. $\Delta$ denotes an improvement over the few-shot fine-tuning method compared to our framework. All models (except fine-tuning with full data) are trained with 10-shot labeled samples for each class and overall F1 aggregated over five different runs with different random seeds.}
\label{tab:main-result}
\end{table*}

\begin{table}
\centering
\resizebox{\linewidth}{!}{
\begin{small}
\begin{tabular}{l | ccc}
\toprule
\bf Models &\bf  SNIPS &\bf  MIT Movie &\bf  CONLL-03 \\
\midrule
SST & 81.07\small{\textpm0.40} & 72.25\small{\textpm0.11} & 72.86\small{\textpm0.34} \\
\textbf{\model} & \bf 87.33\small{\textpm0.30} & \bf 77.73\small{\textpm0.22} & \bf 79.10\small{\textpm0.27} \\
\midrule
\quad w/o. selection  & 82.37\small{\textpm0.39} & 70.12\small{\textpm0.29} & 73.40 \small{\textpm0.37} \\
\quad w/o. confidence & 86.98\small{\textpm0.33} & 76.99\small{\textpm0.29} & 77.60 \small{\textpm0.29} \\
\quad w/o. certainty & 83.33\small{\textpm0.30} & 72.29\small{\textpm0.33} & 73.16\small{\textpm0.31} \\
\quad w/o. PHCE loss & 87.01\small{\textpm0.34} & 77.32\small{\textpm0.27} & 78.31\small{\textpm0.35} \\
\quad w/o. GCR & 87.11\small{\textpm0.31} & 77.25\small{\textpm0.23} & 77.93\small{\textpm0.31} \\
\bottomrule
\end{tabular}
\end{small}
}
\caption{The ablation study results over 10-shot labeled data per class (F1 score \%).}
\label{tab:ablation}
\end{table}

\subsection{Main Results}
Table~\ref{tab:main-result} illustrates the main results of our framework~{\model} compared with other baselines.
The results of fine-tuning over full data are the ceiling performance. 
With only 10 labeled training data per class, we achieve the best averaged F1 score of 77.78\%. In addition, {\model} outperforms the few-shot fine-tuning and standard self-training by 7.99\% and 6.36\%, respectively. 
The vanilla fine-tuning method without any unlabeled data performs worst than expected since the number of labeled data is extremely insufficient for the parameter modification.
The performance of the self-training-based approaches (including SST, MetaST, and SeqUST) is consistently better than others, which indicates the effectiveness of the self-training paradigm.
{MetaST} is a strong baseline that performs a token-level re-weighting strategy in self-training. Yet, {MetaST} ignores the robust learning of some noisy labels even though they have higher weights, and does not consider the consistency regularization, its performance can be considered sub-optimal in contrast with ours.
Results show that {\model} achieves high improvement over the state-of-the-art {MetaST} by 1.25\%.

\subsection{Ablation Study}
In Table~\ref{tab:ablation}, we randomly choose three datasets to conduct an ablation study to investigate the characteristics of the main components in~{\model}. 
The results show that no matter which module is removed, the model performance is affected. All variants still outperform standard self-training, even though removing some components.
When removing the reliable token selection (w/o. selection), the performance declines a lot because that many noisy labels hurt the model effectiveness. 
We also observe that the result of w/o. certainty is lower than w/o. confidence, which demonstrates that the model uncertainty is more useful for label denoising, and greatly alleviate the conformation bias issue.
Moreover, the use of robust loss (i.e., the PHCE loss in MSL) and consistency regularization consistently contribute to the robustness improvement when training on pseudo-labeled data.

\begin{table}[t]
\centering
\resizebox{\linewidth}{!}{
\begin{tabular}{l | ccc | ccc}
\toprule
& \multicolumn{3}{c|}{\textbf{SeqUST}} & \multicolumn{3}{c}{\textbf{MetaST}} \\
\bf \#-shot$\longrightarrow$ &\bf  20 &\bf  50 &\bf 100 &\bf  20 &\bf  50 &\bf  100 \\
\midrule
SNIPS  & 92.13 & 93.44 & 95.60 & 91.99 & 92.82 & 95.10 \\
Multiwoz & 79.70 & 82.05 & 83.36 & 79.45 & 81.34 & 84.27 \\
MIT Movie & 80.80 & 83.16 & 84.98 & 80.29 & 82.75 & 84.35 \\
MIT Restaurant  & 69.02 & 73.95 & 75.70 & 67.93 & 72.83 & 75.28 \\
CoNLL-03 & 81.74 & 83.20 & 85.59 & 78.54 & 82.34 & 85.10 \\
OntoNotes & 82.26 & 84.00 & 85.77 & 80.78 & 82.44 & 85.01 \\
\bottomrule
\end{tabular}
}
\caption{The F1 score with different numbers (20/50/100 examples per class) of labeled data.}
\label{tab:data_efficiency}
\end{table}

\begin{table}[t]
\centering
\resizebox{\linewidth}{!}{
\begin{small}
\begin{tabular}{l | ccc}
\toprule
\bf Selection Strategy &\bf SNIPS &\bf  MIT Movie &\bf  CONLL-03 \\
\midrule
None  & 45.75 & 51.77 & 49.88 \\
Confidence & 43.18 & 48.06 & 45.22 \\
Certainty & 20.42 & 27.60 & 29.10 \\
\bf Confidence + Certainty & \bf 17.76 & \bf  23.10 & \bf  24.63 \\
\bottomrule
\end{tabular}
\end{small}
}
\caption{The error rate (\%) of different selection strategy. None means does not select tokens.}
\label{tab:selection-noise-rate}
\end{table}
\vspace{-.25em}

\begin{figure*}[t]
\centering
\includegraphics[width=0.95\linewidth]{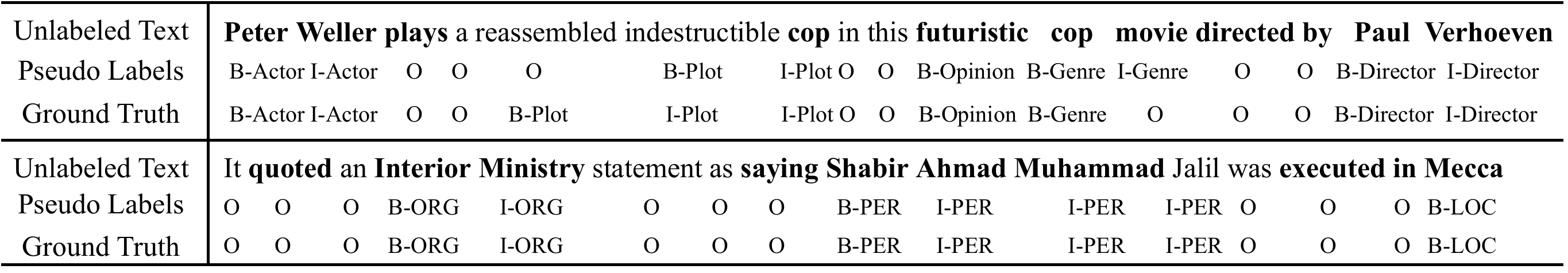}
\caption{Cases of pseudo-labeled texts. The language tokens in \textbf{bold} are sampled as reliable tokens.}.
\label{fig:cases}
\end{figure*}

\subsection{Further Analyses}
\noindent\textbf{Data Efficiency.}
We further explore the model effects with different numbers of labeled data per class (\#-shot) among $\{20, 50, 100\}$. 
Results in Table~\ref{tab:data_efficiency} illustrate that the performance gradually improves as the number of labeled data increases, as expected. 
In addition, we also find that our~{\model} consistently outperforms the strong baseline~{MetaST} no matter how many labeled training examples. This can be attributed to the introduction of uncertainty-aware self-training with well-designed robust learning approaches.

\begin{figure}
\centering
\begin{tabular}{cc}
\begin{minipage}[t]{0.326\linewidth}
    \includegraphics[width = 1\linewidth]{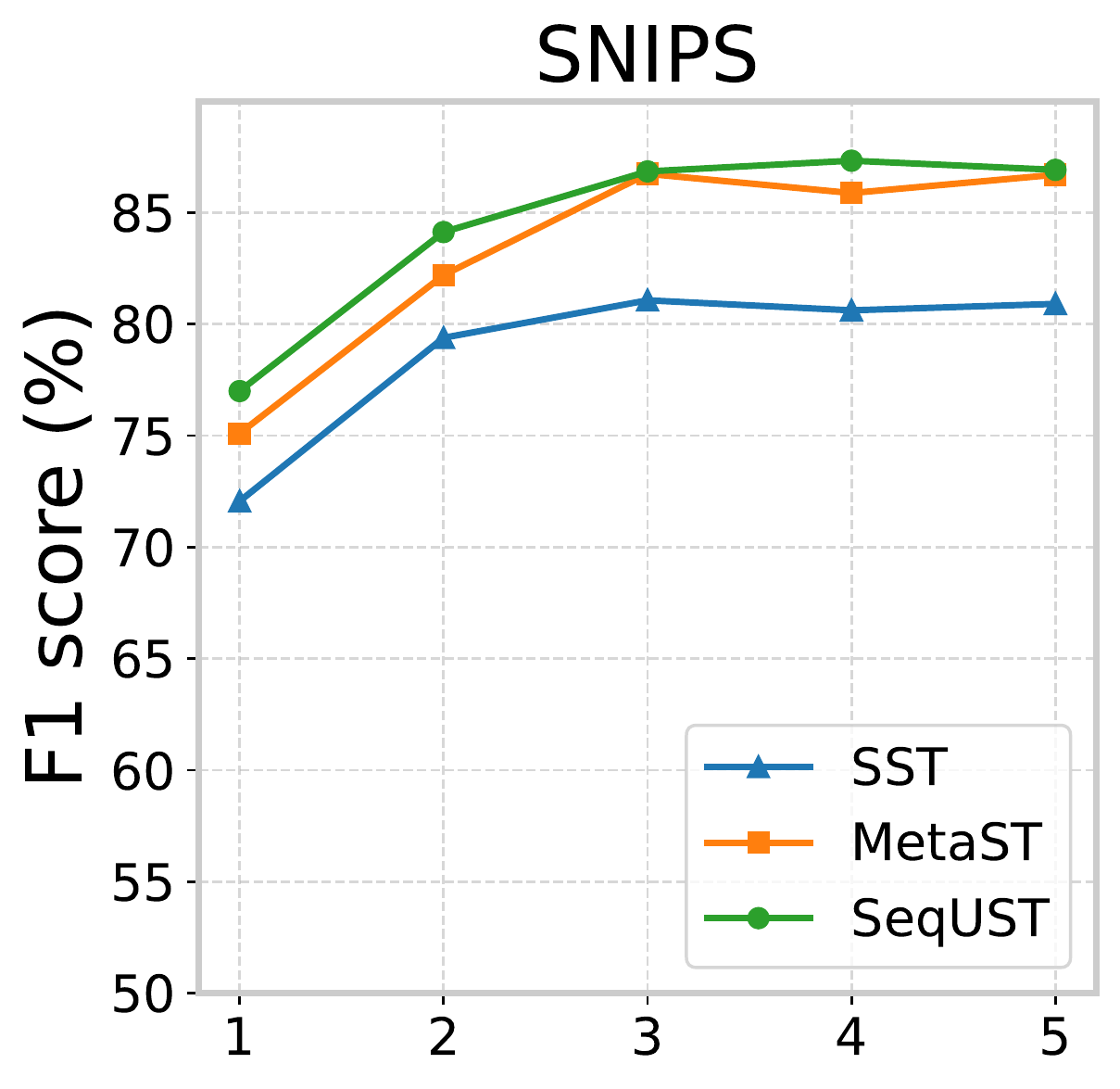}
\end{minipage}
\begin{minipage}[t]{0.3\linewidth}
    \includegraphics[width = 1\linewidth]{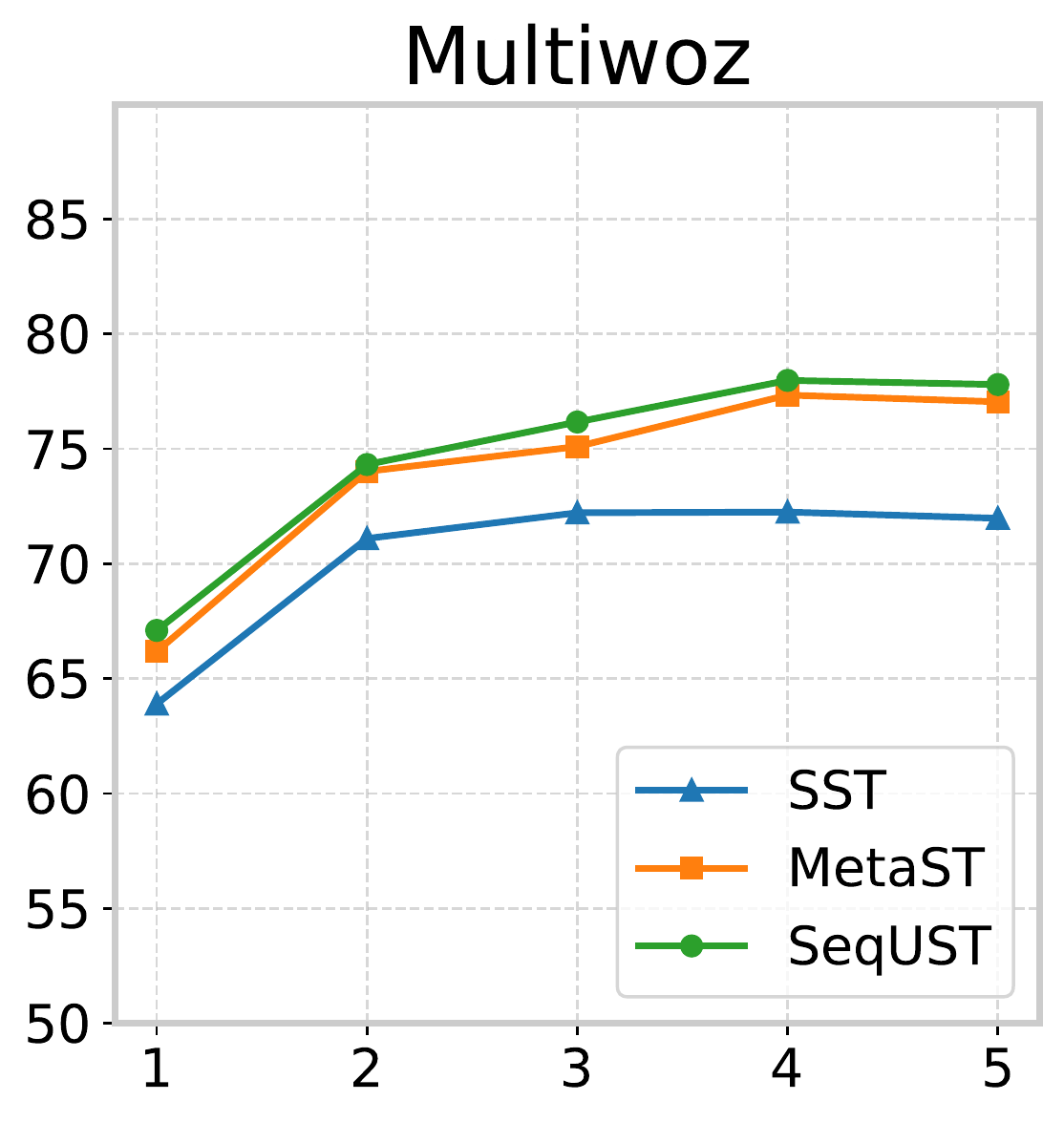}
\end{minipage}
\begin{minipage}[t]{0.3\linewidth}
    \includegraphics[width = 1\linewidth]{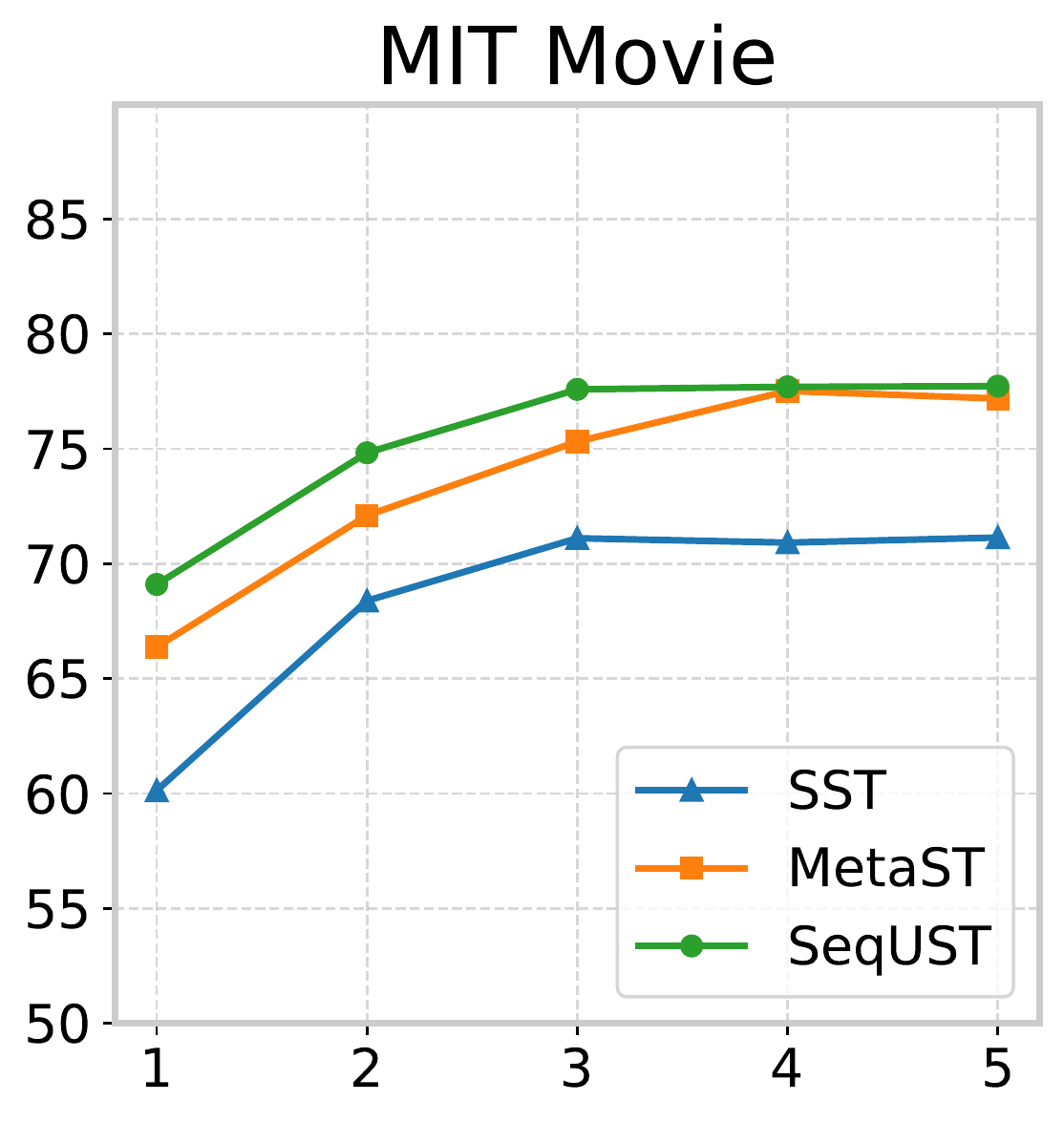}
\end{minipage}
\end{tabular}
\begin{tabular}{cc}
\begin{minipage}[t]{0.326\linewidth}
    \includegraphics[width = 1\linewidth]{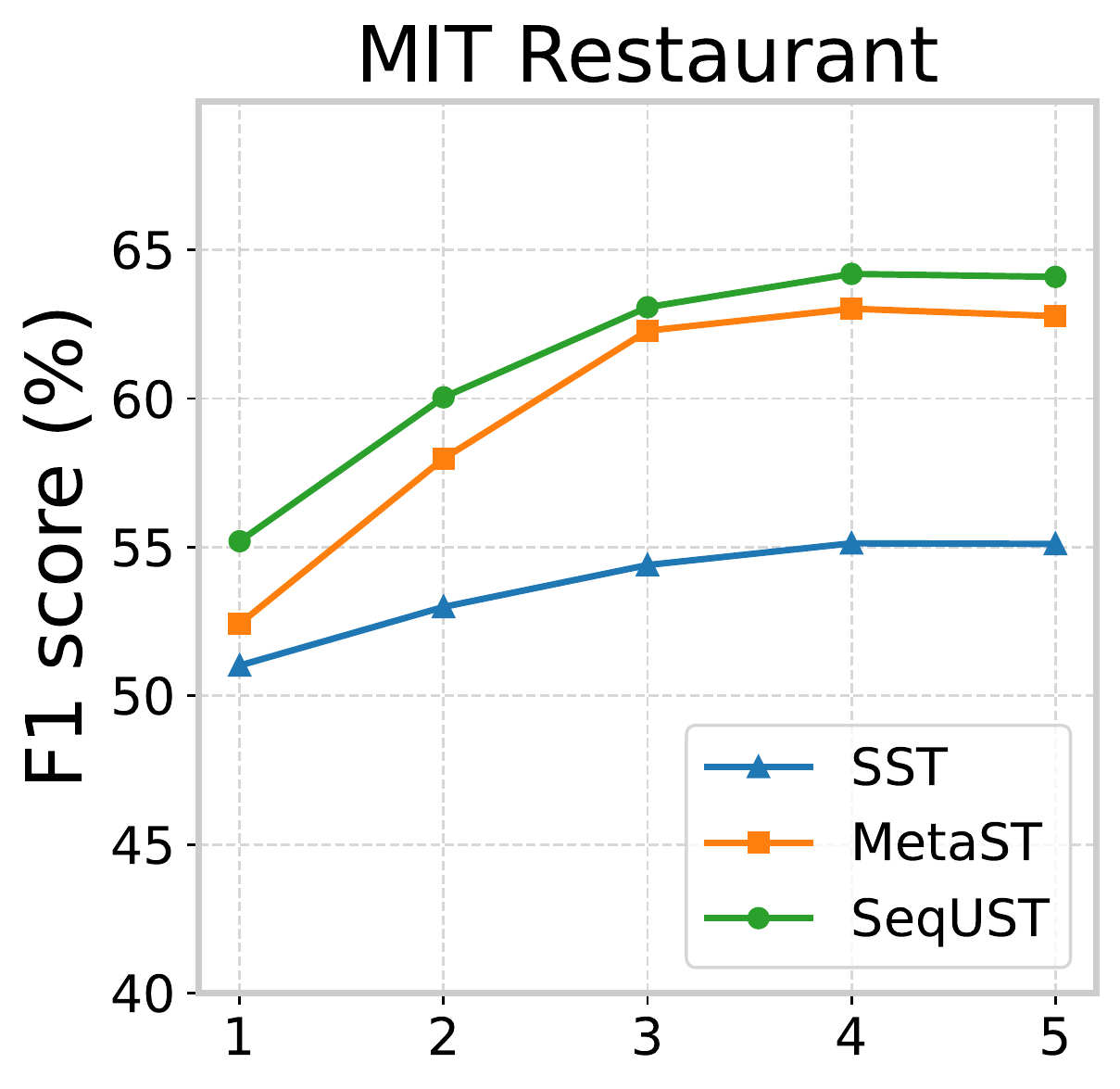}
\end{minipage}
\begin{minipage}[t]{0.3\linewidth}
    \includegraphics[width = 1\linewidth]{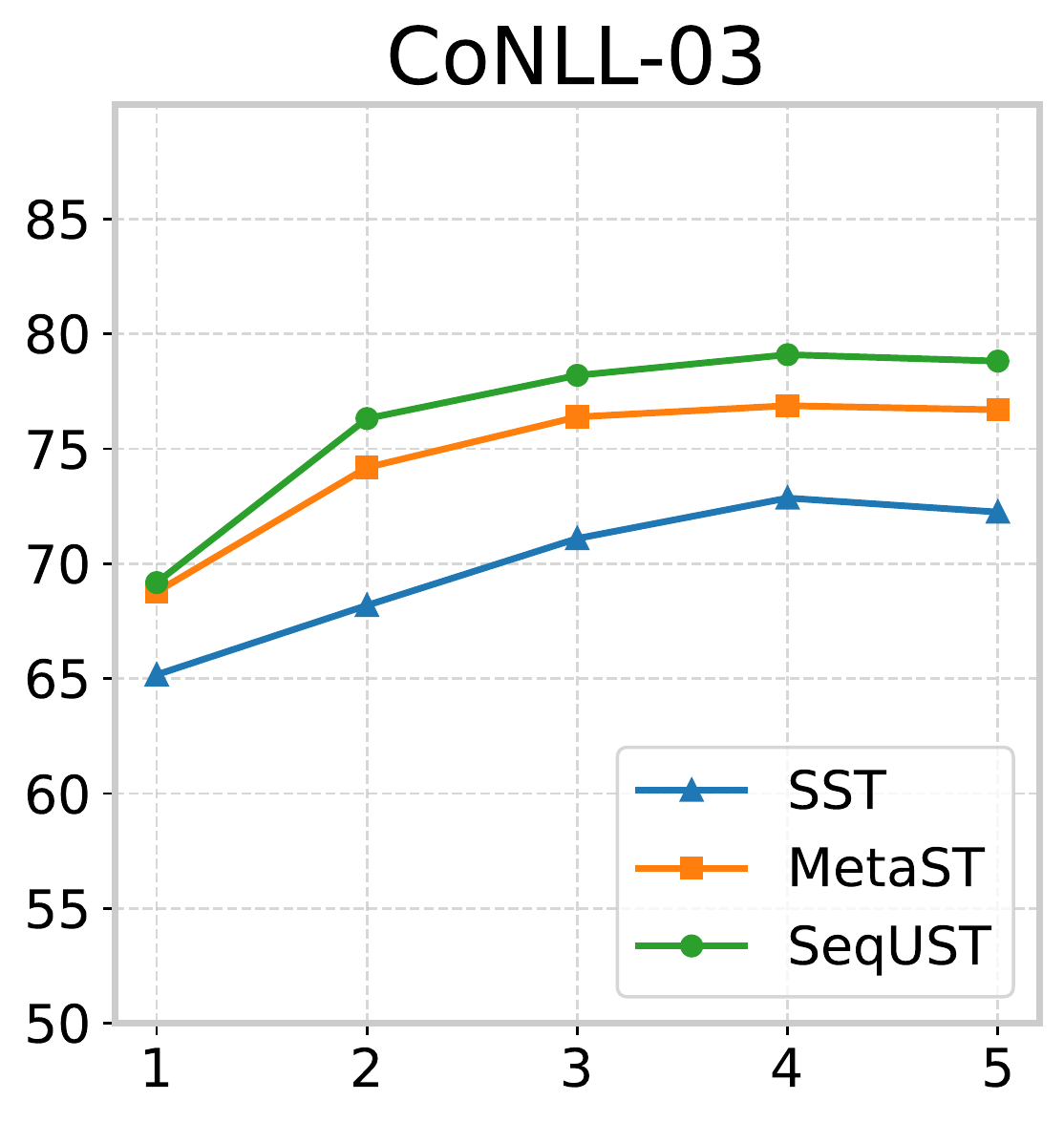}
\end{minipage}
\begin{minipage}[t]{0.3\linewidth}
    \includegraphics[width = 1\linewidth]{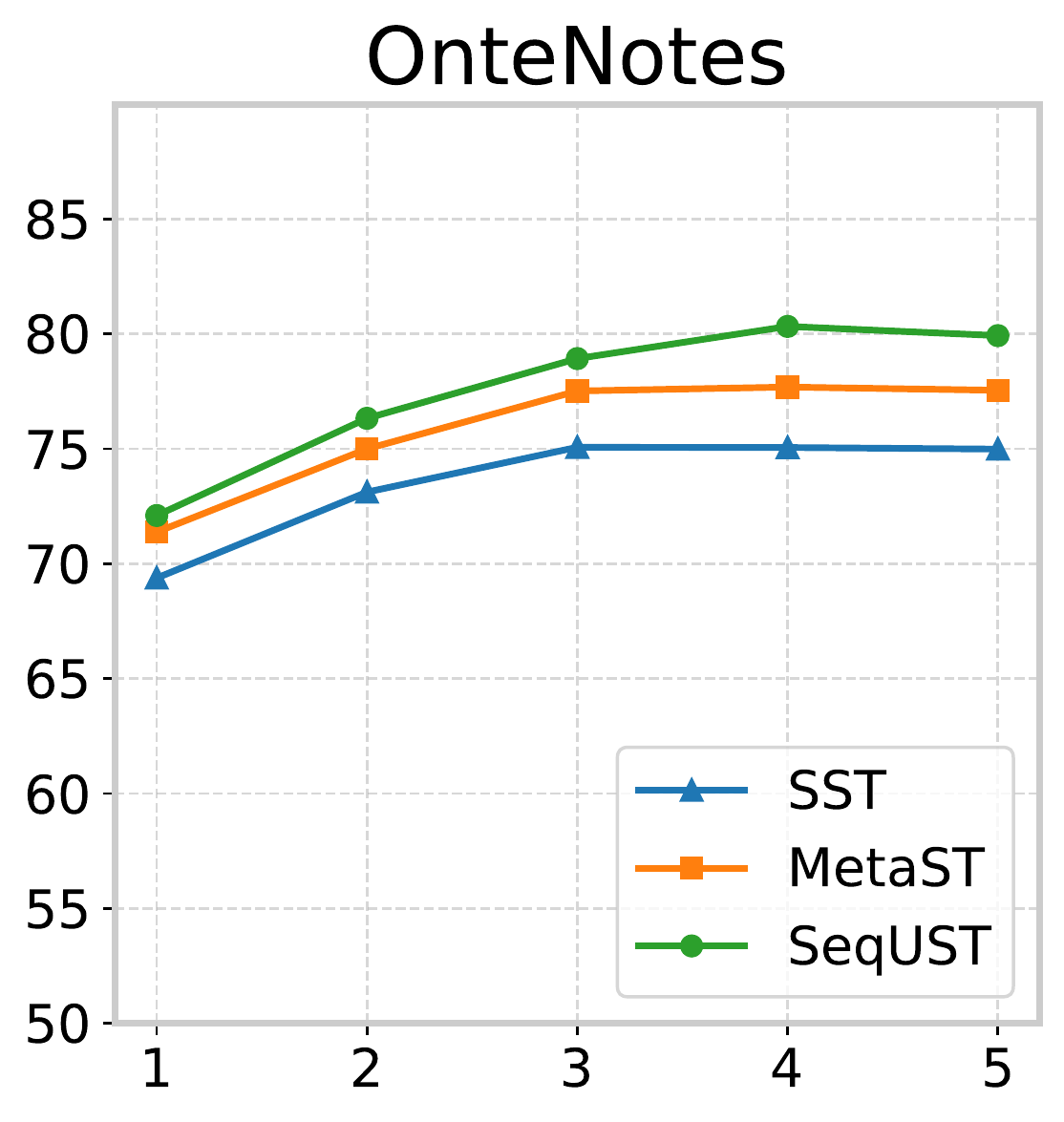}
\end{minipage}
\end{tabular}
\caption{The performance (F1 score \%) of different self-training iterations over six benchmarks.}
\label{fig:robust}
\vspace{-.25em}
\end{figure}

\noindent\textbf{Effectiveness of Reliable Token Selection.}
To demonstrate the effectiveness of reliable token selection, we choose three datasets to investigate the error rate on the unlabeled data. As shown in Table~\ref{tab:selection-noise-rate}, we obtain the lowest error rate with both token-level confidence and certainty, indicating the merit of discarding noisy pseudo labels.
In addition, we respectively choose two cases from MIT Movie and CONLL-03 to show the pseudo labels and selected tokens. The results in Fig.~\ref{fig:cases} show that 1) most tokens can be correctly predicted and 2) our selection strategy can identify some noisy labels.

\begin{figure}[t]
\centering
\begin{tabular}{cc}
\begin{minipage}[t]{0.4\linewidth}
    \includegraphics[width = 1\linewidth]{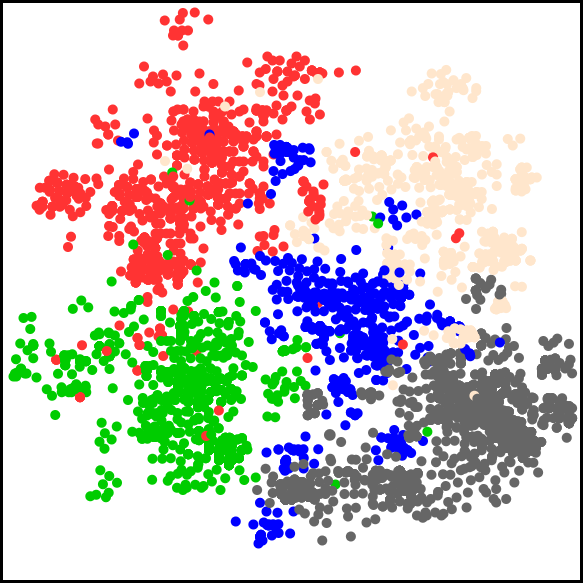}
\end{minipage}
\begin{minipage}[t]{0.4\linewidth}
    \includegraphics[width = 1\linewidth]{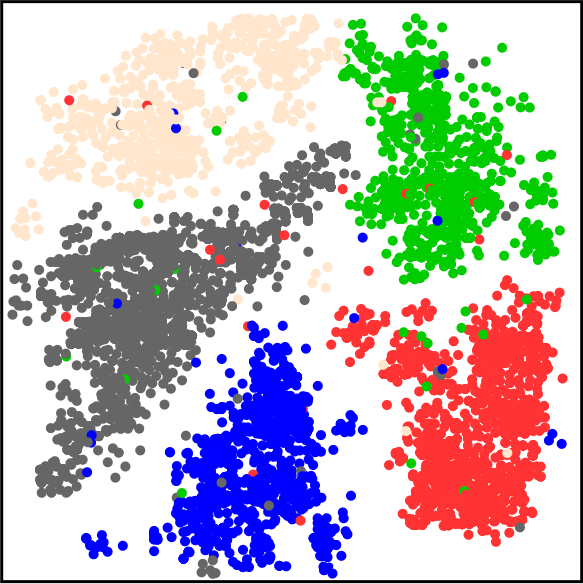}
\end{minipage}
\end{tabular}
\caption{The t-SNE visualization of~{\model} w/o. GCR (left) and w/ GCR (right).}
\label{fig:visualization}
\end{figure}

\noindent\textbf{Effectiveness of the Robust Learning.}
We first demonstrate the importance of PHCE loss for robust learning in the masked sequence labeling task. 
As shown in Fig.~\ref{fig:robust}, our framework consistently outperforms the standard self-training by a large margin, which does not use robust learning methods.
Moreover, the performance of~{MetaST} is not higher than ours, indicating that the model trained by the explicit noise masking and the PHCE loss is more robust than the re-weighting strategy.

In Fig.~\ref{fig:visualization}, we choose the CoNLL-03 testing set and use t-SNE~\cite{van2008visualizing} tool to demonstrate the token representations in the semantic space.
The model trained with Gaussian-based consistency regularization can make a clearer boundary between every two classes,  corroborating our conclusions that avoiding the over-fitting problem and yielding better generalization capability.

\begin{table}
\centering
\resizebox{\linewidth}{!}{
\begin{tabular}{l | cc | cc}
\toprule
\multirow{2}{*}{\textbf{Datasets}} &  \multicolumn{2}{c|}{\textbf{BERT}}  &  \multicolumn{2}{c}{\textbf{BiLSTM}} \\
& SoftMax & CRF & SoftMax & CRF \\
\midrule
SNIPS  & 87.33\small{\textpm0.30} & 86.95\small{\textpm0.25} & 84.94\small{\textpm0.20} & 85.16\small{\textpm0.28} \\
Multiwoz  & 77.98\small{\textpm0.26} & 78.20\small{\textpm0.38} & 73.13\small{\textpm0.19} & 73.92\small{\textpm0.20} \\
MIT Movie & 77.73\small{\textpm0.22} & 78.02\small{\textpm0.24} & 72.05\small{\textpm0.26} & 73.12\small{\textpm0.31} \\
MIT Restaurant & 64.19\small{\textpm0.36} & 66.17\small{\textpm0.26} & 60.83\small{\textpm0.42} & 61.11\small{\textpm0.47} \\
CoNLL-03 & 79.10\small{\textpm0.27} & 78.21\small{\textpm0.35} & 75.53\small{\textpm0.22} & 75.22\small{\textpm0.23} \\
OntoNotes & 80.33\small{\textpm0.44} & 79.74\small{\textpm0.40} & 76.02\small{\textpm0.34} & 76.28\small{\textpm0.34} \\
\midrule
Avg. & 77.78 & 77.88 & 73.75 & 74.14 \\
\bottomrule
\end{tabular}
}
\caption{The F1 score (\%) of different base encoders.}
\label{tab:base-encoder}
\end{table}

\subsection{Performance of Different Base Encoders}
We end this section with a comparison with other base encoders in Table~\ref{tab:base-encoder}, including BERT+SoftMax, BERT+CRF, BiLSTM+SoftMax, and BiLSTM+CRF.
Results show that our framework can be applied to arbitrary encoders.
In addition, CRF is able to exploit the label dependency among the few data and further improves the overall performance of BERT and BiLSTM by 0.10\% and 0.39\%, respectively.

\section{Conclusion}
We propose a novel~{\model} framework for semi-supervised neural sequence labeling based on uncertainty-aware self-training.
We utilize token-level model confidence and certainty to judiciously select reliable tokens in each unlabeled sentence. To mitigates the noisy labels, we introduce a masked sequence labeling task with noise-robust PHCE loss. We also present Gaussian-based consistency regularization to alleviate the over-fitting problem.
Extensive experiments over multiple benchmarks show that our framework consistently outperforms strong baselines in low-resource settings.
In the future, we will further improve the performance of denoising and apply our framework to other token-level scenarios and NLP tasks, such as extractive question answering, event extraction, etc.

\section*{Acknowledgements}

This work was partially supported by the National Natural Science Foundation of China under Grant No. U1911203, 
Alibaba Group through the Alibaba Innovation Research Program, 
the National Natural Science Foundation of China under Grant No. 61877018,
the Research Project of Shanghai Science and Technology Commission (20dz2260300) and The Fundamental Research Funds for the Central Universities.


\bibliography{aaai23}
\appendix

\begin{table}
\centering
\resizebox{\linewidth}{!}{
\begin{tabular}{lc}
\toprule
\bf Hyper-parameter &\bf Value \\
\midrule
Batch Size & \{1, 4, 8, 16\} \\
Learning Rate & \{1e-5, 2e-5, 5e-5, 1e-4, 2e-4\} \\
Dropout Rate & \{0.1, 0.3, 0.5\} \\
$T$ & \{5, 10, 15, 20\} \\
Warmup Rate & \{0.1\} \\
Max Length & \{64\} \\
$\lambda$ & \{0.25, 0.50, 0.75, 1.0\} \\
$K$ & \{1, 2, 3, 4, 5\} \\
$\tau$ & \{10\} \\
\bottomrule
\end{tabular}
}
\caption{The searching space for each hyper-parameter.}
\label{tab:search-scope}
\end{table}

\section{Appendix A: Hyper-parameter Settings}
\label{app:hyper-parameter}
The searching space of each hyper-parameter is shown in Table~\ref{tab:search-scope}. 
Finally, we choose AdamW as the optimizer with a warm-up rate of 0.1. The empirical hyper-parameters are set as $T=20$, $\lambda=0.5$. The appropriate number of perturbed embeddings in Gaussian-based consistency regularization (GCR) is $K=3$.

\section{Appendix B: Main Results with Different Numbers of Labeled Data}
\label{app:main_results}
We perform data efficiency experiments (20-shot, 50-shot and 100-shot) over all datasets for all baselines in Table 8\~10. Results show that our~{\model} consistently outperforms state-of-the-art approaches.

\begin{table*}[t]
\centering
\resizebox{\linewidth}{!}{
\begin{tabular}{lcccccccc}
\toprule
\bf Baselines & \bf  SNIPS & \bf  Multiwoz & \bf  MIT Movie &  \bf  MIT Restaurant & \bf  CoNLL-03 & \bf  OntoNotes & \bf  Avg. & $\Delta$ \\
\midrule
\multicolumn{9}{l}{\textit{\textbf{Full Data}}}\\
Fine-tuning & 97.00 & 88.00 & 87.30 & 79.00 & 91.90 & 89.20 & 88.73 & - \\
\midrule
\multicolumn{9}{l}{\textit{\textbf{Few Labeled Data (20-shot)}}}\\
Fine-tuning & 86.80\textit{\textpm0.38} & 75.33\small{\textpm0.27} & 73.12\small{\textpm0.33} & 55.25\small{\textpm0.25} & 73.89\small{\textpm0.21} & 76.70\small{\textpm0.27} & 73.52 & +0.00 \\
\midrule
\multicolumn{9}{l}{\textit{\textbf{Few Labeled Data (20-shot) + Unlabeled Data}}}\\
SST & 88.37\small{\textpm0.39} & 77.62\small{\textpm0.11} & 75.39\small{\textpm0.22} & 56.03\small{\textpm0.22} & 74.38\small{\textpm0.44} & 78.00\small{\textpm0.30} & 74.97 & +1.45\\
VAT  & 87.03\small{\textpm0.23} & 75.32\small{\textpm0.35} & 75.93\small{\textpm0.29} & 56.70\small{\textpm0.32} & 74.95\small{\textpm0.33} & 78.12\small{\textpm0.29} & 74.68 & +1.16 \\
SeqVAT  & 87.16\small{\textpm0.38} & 75.51\small{\textpm0.31} & 75.96\small{\textpm0.35} & 56.73\small{\textpm0.30} & 74.62\small{\textpm0.29} & 78.33\small{\textpm0.31} & 74.72 & +1.2\\
CVT  & 86.96\small{\textpm0.34} & 75.10\small{\textpm0.22} & 75.48\small{\textpm0.50} & 56.17\small{\textpm0.40} & 74.70\small{\textpm0.53} & 78.18\small{\textpm0.41} & 74.43 & +0.91\\
MetaST  & \underline{91.99\small{\textpm0.23}} & \underline{79.45\small{\textpm0.42}} & \underline{80.29\small{\textpm0.31}} & \underline{67.93\small{\textpm0.34}} & \underline{78.54\small{\textpm0.49}} & \underline{80.87\small{\textpm0.28}} & \underline{79.85} & \underline{+6.33} \\
\midrule
\textbf{SeqUST}  & \bf 92.13\small{\textpm0.22} & \bf 79.70\small{\textpm0.31} & \bf 80.80\small{\textpm0.28} & \bf 69.02\small{\textpm0.35} & \bf 81.74\small{\textpm0.27} & \bf 82.26\small{\textpm0.39} & \bf 80.94 & \bf +7.42 \\
\bottomrule
\end{tabular}
}
\caption{The performance comparison of F1 scores (\%) with standard deviations on six benchmarks (20-shot).}
\label{tab:main-result-20}
\end{table*}

\begin{table*}[t]
\centering
\resizebox{\linewidth}{!}{
\begin{tabular}{lcccccccc}
\toprule
\bf Baselines & \bf  SNIPS & \bf  Multiwoz & \bf  MIT Movie &  \bf  MIT Restaurant & \bf  CoNLL-03 & \bf  OntoNotes & \bf  Avg. & $\Delta$ \\
\midrule
\multicolumn{9}{l}{\textit{\textbf{Full Data}}}\\
Fine-tuning & 97.00 & 88.00 & 87.30 & 79.00 & 91.90 & 89.20 & 88.73 & - \\
\midrule
\multicolumn{9}{l}{\textit{\textbf{Few Labeled Data (50-shot)}}}\\
Fine-tuning & 89.36\textit{\textpm0.38} & 77.17\small{\textpm0.11} & 76.01\small{\textpm0.25} & 58.83\small{\textpm0.23} & 75.20\small{\textpm0.23} & 78.16\small{\textpm0.29} & 75.79 & +0.00 \\
\midrule
\multicolumn{9}{l}{\textit{\textbf{Few Labeled Data (50-shot) + Unlabeled Data}}}\\
SST & 89.95\small{\textpm0.22} & 78.14\small{\textpm0.34} & 77.82\small{\textpm0.29} & 60.80\small{\textpm0.23} & 77.16\small{\textpm0.30} & 79.13\small{\textpm0.30} & 77.15 & +1.36\\
VAT  & 88.80\small{\textpm0.21} & 77.93\small{\textpm0.30} & 77.99\small{\textpm0.29} & 61.25\small{\textpm0.33} & 78.32\small{\textpm0.33} & 79.57\small{\textpm0.20} & 77.31 & +1.52 \\
SeqVAT  & 88.92\small{\textpm0.34} & 77.88\small{\textpm0.30} & 78.21\small{\textpm0.27} & 61.22\small{\textpm0.19} & 78.82\small{\textpm0.28} & 79.66\small{\textpm0.32} & 77.45 & +1.66\\
CVT  & 88.02\small{\textpm0.44} & 77.77\small{\textpm0.38} & 77.94\small{\textpm0.40} & 62.08\small{\textpm0.42} & 77.82\small{\textpm0.35} & 79.30\small{\textpm0.29} & 75.83 & +0.04\\
MetaST  & \underline{92.82\small{\textpm0.23}} & \underline{81.34\small{\textpm0.39}} & \underline{82.75\small{\textpm0.29}} & \underline{72.83\small{\textpm0.22}} & \underline{82.34\small{\textpm0.32}} & \underline{82.44\small{\textpm0.43}} & \underline{82.42} & \underline{+6.63} \\
\midrule
\textbf{SeqUST}  & \bf 93.44\small{\textpm0.31} & \bf 82.05\small{\textpm0.24} & \bf 83.16\small{\textpm0.20} & \bf 73.95\small{\textpm0.30} & \bf 83.20\small{\textpm0.29} & \bf 84.00\small{\textpm0.47} & \bf 83.3 & \bf +7.51 \\
\bottomrule
\end{tabular}
}
\caption{The performance comparison of F1 scores (\%) with standard deviations on six benchmarks (50-shot).}
\label{tab:main-result-50}
\end{table*}

\begin{table*}[t]
\centering
\resizebox{\linewidth}{!}{
\begin{tabular}{lcccccccc}
\toprule
\bf Baselines & \bf  SNIPS & \bf  Multiwoz & \bf  MIT Movie &  \bf  MIT Restaurant & \bf  CoNLL-03 & \bf  OntoNotes & \bf  Avg. & $\Delta$ \\
\midrule
\multicolumn{9}{l}{\textit{\textbf{Full Data}}}\\
Fine-tuning & 97.00 & 88.00 & 87.30 & 79.00 & 91.90 & 89.20 & 88.73 & - \\
\midrule
\multicolumn{9}{l}{\textit{\textbf{Few Labeled Data (100-shot)}}}\\
Fine-tuning & 91.10\textit{\textpm0.18} & 78.38\small{\textpm0.25} & 78.89\small{\textpm0.23} & 62.07\small{\textpm0.30} & 77.44\small{\textpm0.28} & 78.29\small{\textpm0.29} & 77.70 & +0.00 \\
\midrule
\multicolumn{9}{l}{\textit{\textbf{Few Labeled Data (100-shot) + Unlabeled Data}}}\\
SST & 92.70\small{\textpm0.29} & 80.27\small{\textpm0.31} & 80.02\small{\textpm0.32} & 66.12\small{\textpm0.24} & 81.03\small{\textpm0.30} & 82.22\small{\textpm0.26} & 80.39 & +2.69\\
VAT  & 91.88\small{\textpm0.32} & 80.11\small{\textpm0.24} & 80.06\small{\textpm0.31} & 66.89\small{\textpm0.39} & 81.74\small{\textpm0.27} & 83.07\small{\textpm0.25} & 80.63 & +2.93 \\
SeqVAT  & 92.30\small{\textpm0.39} & 80.09\small{\textpm0.32} & 80.12\small{\textpm0.35} & 66.66\small{\textpm0.33} & 80.93\small{\textpm0.27} & 82.97\small{\textpm0.29} & 80.51 & +2.81\\
CVT  & 92.01\small{\textpm0.34} & 80.30\small{\textpm0.30} & 79.87\small{\textpm0.21} & 66.40\small{\textpm0.42} & 81.41\small{\textpm0.33} & 83.10\small{\textpm0.41} & 80.52 & +2.82\\
MetaST  & \underline{95.10\small{\textpm0.33}} & \underline{84.27\small{\textpm0.51}} & \underline{84.35\small{\textpm0.39}} & \underline{75.28\small{\textpm0.29}} & \underline{85.10\small{\textpm0.41}} & \underline{85.01\small{\textpm0.24}} & \underline{84.85} & \underline{+7.15} \\
\midrule
\textbf{SeqUST}  & \bf 95.60\small{\textpm0.29} & \bf 84.36\small{\textpm0.28} & \bf 84.48\small{\textpm0.31} & \bf 75.70\small{\textpm0.35} & \bf 85.59\small{\textpm0.25} & \bf 85.27\small{\textpm0.40} & \bf 85.17 & \bf +7.47 \\
\bottomrule
\end{tabular}
}
\caption{The performance comparison of F1 scores (\%) with standard deviations on six benchmarks (100-shot).}
\label{tab:main-result-100}
\end{table*}

\end{document}